\documentclass[review]{elsarticle}
\graphicspath{ {./figures/} }
\usepackage{hyperref}
\usepackage{float}
\usepackage{tabulary,xcolor}
\usepackage{graphicx,times,amsmath}
\usepackage{blindtext}
\usepackage{amsmath}
\usepackage{amsfonts,amsmath,amssymb}
\usepackage{booktabs}

\usepackage{adjustbox}
\usepackage{array}
\usepackage[caption=false, font=footnotesize]{subfig}
\usepackage{bm}
\usepackage{enumerate}
\usepackage{fixltx2e}

\usepackage{caption}

\usepackage{verbatim} %comments
\usepackage{apalike}
\restylefloat{figure}
\restylefloat{table}
\usepackage{listings,xcolor}

\lstdefinestyle{listing_style}{frame=single,basicstyle=\fontfamily{pcr}\selectfont,numberstyle=\tiny,xleftmargin=1pc,linewidth=.98\linewidth,backgroundcolor=\color{black!0},breaklines=true,keywordstyle=\color{blue},commentstyle=\color{darkgray},numbers=left,tabsize=3,captionpos=b,escapeinside={[@}{@]}}

\journal{Expert Systems with Applications}

\biboptions{authoryear}

\begin{document}
\begin{frontmatter}

\begin{titlepage}
\begin{center}
\vspace*{1cm}

\textbf{ \large Novel Multicolumn Kernel Extreme Learning Machine for Food Detection via Optimal Features from CNN}

\vspace{1.5cm}

% Author names and affiliations
Ghalib Ahmed Tahir$^{a}$ (ghalib@siswa.um.edu.my), Chu Kiong Loo$^a$ (ckloo.um@edu.my) \\

\hspace{10pt}

\begin{flushleft}
\small  
$^a$ Department of Artificial Intelligence, Faculty Of Computer Science And Information Technology, University of Malaya, 50603 Kuala Lumpur, Malaysia \\ 
% $^b$ Department of Artificial Intelligence, Faculty Of Computer Science And Information Technology, University of Malaya, 50603 Kuala Lumpur, Malaysia  \\

\begin{comment}
Clearly indicate who will handle correspondence at all stages of refereeing and publication, also post-publication. Ensure that phone numbers (with country and area code) are provided in addition to the e-mail address and the complete postal address. Contact details must be kept up to date by the corresponding author.
\end{comment}

\vspace{1cm}
\textbf{Corresponding Author:} \\
Chu Kiong Loo \\
Department of Artificial Intelligence, Faculty Of Computer Science And Information Technology, \\
	University of Malaya, 50603 Kuala Lumpur, Malaysia  \\
Email: ckloo.um@um.edu.my

\end{flushleft}        
\end{center}
\end{titlepage}

\title{Novel Multicolumn Kernel Extreme Learning Machine for Food Detection via Optimal Features from CNN}

\author[label1]{Ghalib Ahmed Tahir}
\ead{ghalib@siswa.um.edu.my}

\author[label1]{Chu Kiong Loo}
\ead{ckloo.um@um.edu.my}

\cortext[cor1]{Corresponding author.}
\address[label1]{Department of Artificial Intelligence, Faculty Of Computer Science And Information Technology, University of Malaya, 50603 Kuala Lumpur, Malaysia}
% \address[label2]{Full address of second author, including the country name}
% \address[label3]{Full address of last author, including the country name}

\begin{abstract}
Automatic food detection is an emerging topic of interest due to its wide array of applications ranging from detecting food images on social media platforms to filtering non-food photos from the users in dietary assessment apps. Recently, during the COVID-19 pandemic, it has facilitated enforcing an eating ban by automatically detecting eating activities from cameras in public places. Therefore, to tackle the challenge of recognizing food images with high accuracy, we proposed the idea of a hybrid framework for extracting and selecting optimal features from an efficient neural network. There on, a nonlinear classifier is employed to discriminate between linearly inseparable feature vectors with great precision. In line with this idea, our method extracts features from MobileNetV3, selects an optimal subset of attributes by using Shapley Additive exPlanations (SHAP) values, and exploits kernel extreme learning machine (KELM) due to its nonlinear decision boundary and good generalization ability. However, KELM suffers from the 'curse of dimensionality problem' for large datasets due to the complex computation of kernel matrix with large numbers of hidden nodes. We solved this problem by proposing a novel multicolumn kernel extreme learning machine (MCKELM) which exploited the k-d tree algorithm to divide data into N subsets and trains separate KELM on each subset of data. Then, the method incorporates KELM classifiers into parallel structures and selects the top k nearest subsets during testing by using the k-d tree search for classifying input instead of the whole network. For evaluating a proposed framework large food/non-food dataset is prepared using nine publically available datasets. Experimental results showed the superiority of our method on an integrated set of measures while solving the problem of 'curse of dimensionality in KELM  for large datasets. 
\end{abstract}

\begin{keyword}
Multicolumn\sep Kernel Extreme Learning Machine\sep MobileNet\sep Food Detection\sep Explainable AI\sep SHAP
\end{keyword}

\end{frontmatter}

\section{Introduction}
\begin{figure}[!htbp]
\centering \makeatletter\IfFileExists{images/PipelineFood-NonFoodRecognitionn.png}{\includegraphics[width=4in]{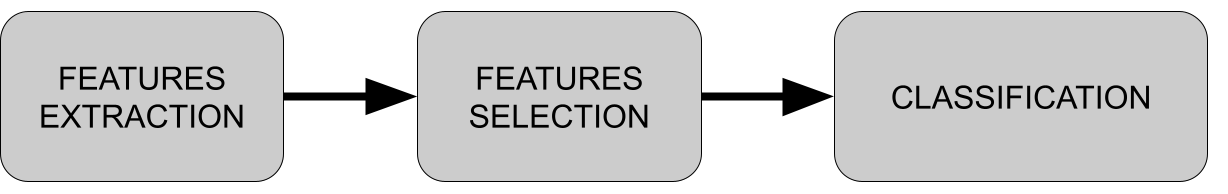}}{}
\makeatother 
\caption{{Shows general pipeline of food detection}}
\label{f-a2619733bd92}
\end{figure}

Automatic detection of food images applications includes visual-based dietary assessment and detecting eating activities from wearable camera photos. The visual-based dietary assessment method reduces the burden of manually collecting the food data by helping users in refreshing their memory using the food pictures of the previous dietary intake. Filtering of non-food images from users is an essential step in these mHealth apps to ensure relevant images are analyzed. 
\mbox{}\protect\newline Similarly, food detection from photos of a wearable camera of the surrounding environment tackles malnutrition in low-income countries and helps to facilitate the process of accessing food intake. Besides this, filtering the non-food images from food images in the user profile on social media fasten the process of analyzing dietary preferences, with minimum human intervention. Visual-based methods for food detection provide alerts to users that are bringing food items to restricted areas where all or certain items are not allowed. One of the related use-case is to ensure the enforcement of a 'no food policy' at the workplaces due to safety concerns. The recent pandemic has also given birth to several use cases of food detection, especially in improving the effectiveness of enforcing the ban on eating in public places or transport to combat the spread of the COVID-19 virus.
Despite the significant importance of this area, there involved research challenges in preparing large datasets and novel machine learning methods, which should be solved in a highly accurate framework of food detection.
Figure~\ref{f-a2619733bd92} illustrates the basic steps in a pipeline of food recognition.

In line with this idea, this study employed an efficient convolutional neural network for real-time feature extraction and then selected optimal features from high dimensional feature space using SHAP activation values from the gradient explainer. Finally, this work contributed mainly to the classification phase of the pipeline by proposing a novel algorithm to address challenges with existing methods, as recent approaches are using convolutional neural networks, both for feature representations of the images and classification. However, linear classifiers in a convolutional neural network (CNN) do not suit linearly inseparable features. Consequently, requiring machine learning algorithms that discriminate linearly inseparable feature vectors with a high degree of accuracy, while achieving the objective of real-time classification in a distributed cloud environment.

 Kernel extreme learning machine is one of the advanced methods in extreme learning machines for classifying linearly inseparable features, and it has good generalization ability. However, KELM relies on complex inner matrix computations to cater to large data volumes, making these operations costlier in terms of computational resources and processing time. Similarly, unlike KELM, which has a strongly coupled network structure, preferred machine learning algorithms take advantage of the cloud's parallel and distributed environments. There are certain efforts for other variants of neural networks, showing improved results and performance in a distributed environment like a parallel-structured ANN. \unskip~\cite{1092609:22126340}\unskip~\cite{1092609:22126341} . Although currently implemented strategies in these methods are naive, this paves the pathway for the researchers to develop advanced methods that take advantage of the distributed cloud environment while having competitive performance.  \mbox{}\protect\newline One of the possible strategies to solve this challenge is decoupling strongly coupled neural networks into loosely coupled networks. We achieve this objective by using a space partitioning data structure 'k-d tree' that divides the data into equal distributions. This approach makes it very efficient for nearest neighbor search, and it ensures the equal distribution of data by dividing the data into smaller subsets. Using the k-d tree for dividing data into equal subsets in KELM seems promising as two fundamental problems are resolved in KELM making them suitable for large vision-based datasets 1) Curse of dimensionality and 2) Strongly coupled network structure. \mbox{}\protect\newline This novel classifier is named as multicolumn kernel extreme learning machine (MCKELM) classifier as it solves the dimensionality problem and takes advantage of a distributed cloud-based environment by decoupling the network structure of KELM during the classification stage of food detection. The proposed classifier uses reliable features from the efficient neural network after filtering irrelevant features using feature selection methods that use SHAP values from Gradient Explainer.  Lastly, the visualizations by GradientExplainer highlight the pixels in the photo based on SHAP values, which helps to understand the contribution of every region in the image. Hence improving the interpretability of the proposed architecture compared to existing techniques during the feature selection process. 
\par In summary, the contribution of this paper is summarized as follows.
\par 1) Prepared a large dataset for food detection using nine publically available datasets.
\par 2) Exploited efficient neural networks for real-time feature extraction 
\par 3) Employed feature selection strategy for reliable feature selection using SHAP values from gradient explainer 
\par 4) Proposed novel MCKELM by successfully integrating k-d tree with KELM to solve the curse of dimensionality problem.
 \par We organized our work as follows. The relevant studies of Food detection are discussed in section 2, followed by a detailed explanation of our proposed approach in section 3. In section 4, we discussed the evaluation criteria and the results of our proposed methods in the context of food recognition. Finally, in section 5, we discussed the results, followed by a conclusion and future research work in this area.
 
\section{\textbf{Literature Review}}
Recent methods employ CNN for food detection as  CNN consists of convolution layers and pooling layers which learns efficiently high-level features of the data. However, linear-fully connected classifiers in CNN have reduced classification performance for feature vectors that are not linearly inseparable. Similarly, the training process of CNN is sensitive to the local minimum in the training error surface. Therefore, the generalization performance of fully connected layers in CNN is not always strong enough. Moreover, a non-linear classifier that uses visual features via CNN faces a challenge in selecting optimal features from high dimension space. This increases the importance of new efficient hybrid methods for food detection that solves challenges involved in choosing optimal attributes from CNN and reducing the high complexity of non-linear classifiers. This section discusses the existing literature on food detection in context with these challenges. There on summarized the research gap of existing methods followed by the contribution of this work to existing literature.

 \mbox{}\protect\newline   \unskip~\cite{1092609:22126358} have used handcrafted features with a support vector machine for food/non-food classification. The attributes are based on the color histograms, detected image patterns and DCT coefficients, etc. Their method has used handcrafted features and outperformed the present methods that use visual feature vectors via a convolutional neural network.
\mbox{}\protect\newline \unskip~\cite{1092609:22126357} has proposed a CNN network for food/non-food recognition. The neurons in the multilayer neural network process small patches of the previous layers and are robust against small rotations and shifts. They have used a Cuda-Covent c++ implementation of CNN on a graphics processing unit (GPU), and they also prepared a midsize food/non-food dataset for this purpose. However, they have tested their method on a small size food/non-food dataset and had limited generalizability.
 \mbox{}\protect\newline \unskip~\cite{1092609:22126356} have proposed a CNN-NIN network for food detection. The network consists of four convolutional layers with two 'mplconv' layers. They employed the pre-trained model to copy mid-level image representations and model parameters. The dimension of output classes is reduced from 1000 to 2 for the food/non-food classification. The main advantage of their model is memory efficiency. However, they tested against a small food/non-food dataset. \unskip~\cite{1092609:22126355} has used the pre-trained model google-net for food/non-food classification and conducted experiments on their datasets gathered from various online sources. They have achieved an overall accuracy of 99.2\% and 83.6\% for food class recognition.  \mbox{}\protect\newline   Several studies proposed deep learning models to detect food portions from images of egocentric food datasets \unskip~\cite{1092609:22126338} after gathering their datasets from wearable devices using the e-buttons. The major objective of their research is to detect food items from egocentric images. The gathered dataset contains 29515 images from research participants from week-long recordings, and they have achieved an accuracy of 91.5\% for food detection. However,  their method is evaluated on limited data, and secondly, the fully connected layer in CNN can not discriminate features with high precision.  \unskip~\cite{1092609:22126354} proposed composite architecture for the real-time classification of egocentric food/non-food images. It consists of the deep neural network, shallow learning network, and probabilistic network interface. Their combined method performs better than other deep learning approaches based on an integrated set of measures. 
 \mbox{}\protect\newline The extensive review of existing methods showed that the CNN network has attained state-of-the-art performance on various food/non-food data sets. However, there are several research gaps. At first, large datasets do not exist for the food detection problem. Most of the methods are evaluated for small datasets. Secondly, the linear classifier in CNN is less appropriate for linearly inseparable features. To address this, we evolved KELM and proposed MCKELM to solve the 'curse of dimensionality problem faced by a single KELM for a large dataset. MCKELM has a good generalization ability for linearly inseparable features. Moreover, existing frameworks are not explainable in contrast to our framework, which uses Gradient Explainer to highlight the pixels in the photo hence helping in visualizing areas that are given more importance during the feature selection process. The following section briefly describes our methodology.
    
\section{\textbf{Methodology}}
\begin{figure}[!ht]
\centering
\includegraphics[width=6.0in,height=1.6in]{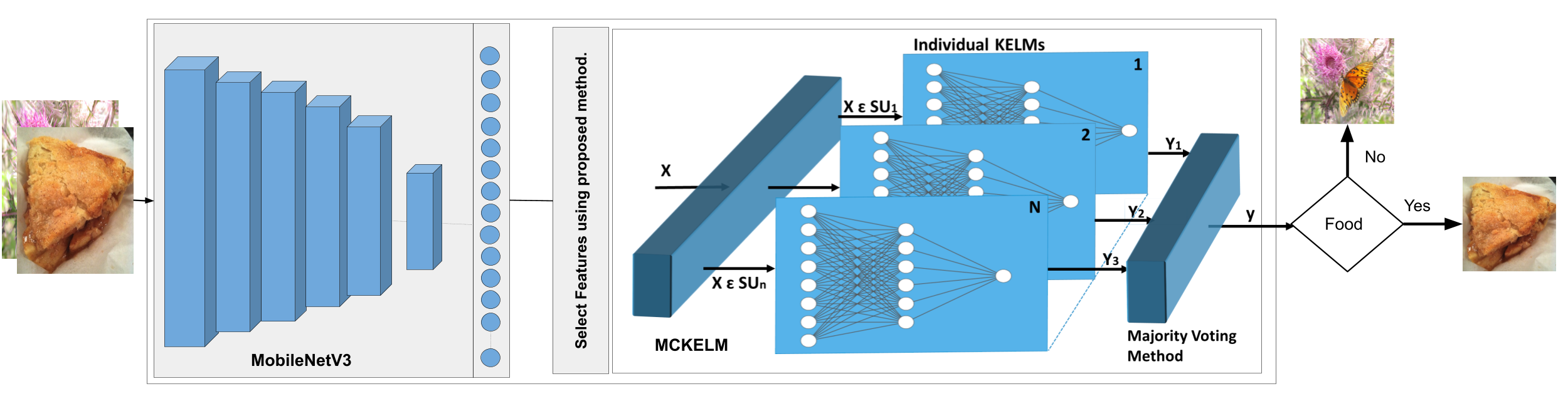}

\caption{General architecture of the proposed framework for food detection.}
\label{f-f1872c66534f}
\end{figure}

The proposed framework extracts the features from a fully connected layer of an efficient convolutional neural network and then selects reliable attributes for reducing complexity. There on, it employs a non-linear classifier for classification. Figure~\ref{f-f1872c66534f}  shows the architecture diagram of the proposed framework. It is to be, noted that we fine-tuned the pre-trained Imagenet model by using transfer learning on our dataset. After feature extraction, we introduced the strategy based on Shapley-value-based interpretations to rank and select the best features. Finally, for the classification phase, we proposed a novel MCKELM. The proposed classifier has good generalization ability and solves the problem faced by KELM for large datasets.

\subsection{\textbf{Efficient Feature Extraction Using MobileNet}} 
\begin{figure}[!ht]
\centering
\includegraphics[width=5.0in]{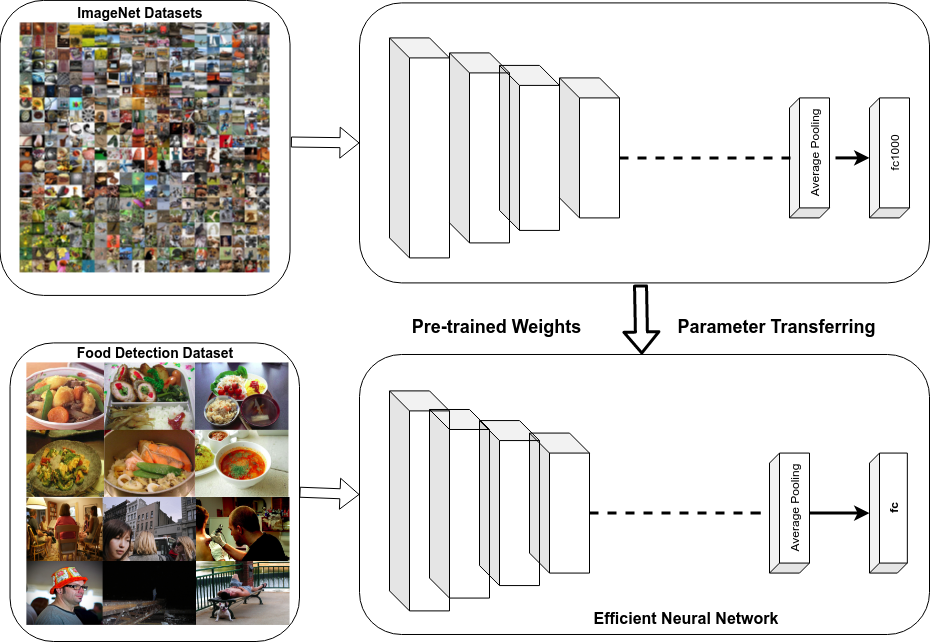}

\caption{Transfer learning to fine-tune pre-trained models on food detection datasets for feature extraction.}
\label{f-finetuning}
\end{figure}

At first, all the variants of MobileNet are finetuned on our food detection dataset. Figure \ref{f-finetuning} shows the finetuning process on the food/non-food detection dataset. There on, fine-tuned model of efficient neural network is employed for real-time features extraction, followed by reliable feature selection and detection of food/non-food classes with high accuracy.

Then we evaluated performance of MobileNetV1\unskip~\cite{1092609:22126331}, MobileNetV2\unskip~\cite{1092609:22126332}, and MobileNetV3\unskip~\cite{1092609:22126333} for feature extraction from food/non-food dataset.  The brief details of MobileNet, including their strength and weakness, are as follows. MobileNetV1 is streamlined architecture and uses depthwise separable convolutions for constructing lightweight deep convolutional neural networks. In it, depthwise convolution performs a single convolution on each of the input channels, and the point convolution filter then linearly combines the output using convolutions. Figure~\ref{f-e9e6685ff449} shows the MobileNetV1 block diagram.
\bgroup

\begin{figure}[!htbp]
\centering \makeatletter\IfFileExists{images/mobilenetv1.png}{\includegraphics[width=4in]{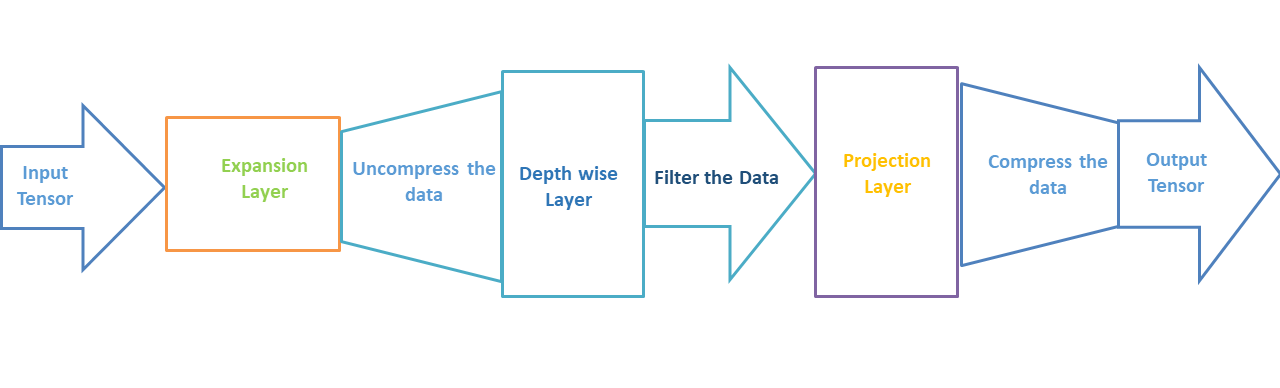}}{}
\makeatother 
\caption{{MobileNetV1 block}}
\label{f-e9e6685ff449}
\end{figure}
\egroup

\begin{figure}[!ht]
\centering
\includegraphics[width=4.0in,height=1.4in]{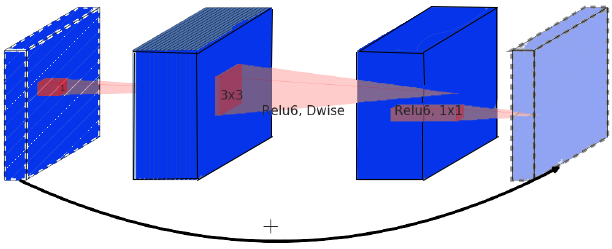}

\caption{MobileNetV2 block with residual}
\label{f-ca45c6253645}
\end{figure}

\mbox{}\protect\newline MobileNetV2 expands over the idea of MobileNetV1 and adds the expansion layer in the block for getting the system of expansion-filtering-compression Figure~\ref{f-ca45c6253645}. Inverted Residual Block introduced in MobileNetV2 improves the performance as each block consists of narrow input and output layer does not have nonlinearity. This approach is followed by higher dimensional space and projection to output. Eventually, the residual connects the bottleneck.

% \bgroup

% \begin{figure}[!htbp]
% \centering \makeatletter\IfFileExists{images/1_baxdp8rs5x_evmnjsd1urg.png}{\includegraphics[width=4in]{images/1_baxdp8rs5x_evmnjsd1urg.png}}{}
% \makeatother 
% \caption{{MobileNetV2 block with residual}}
% \label{f-ca45c6253645}
% \end{figure}
% \egroup

% \bgroup

\begin{figure}[!htbp]
\centering \makeatletter\IfFileExists{images/610c2b51-fb45-4c69-ad6d-27d6906b9cc9.png}{\includegraphics[width=4in]{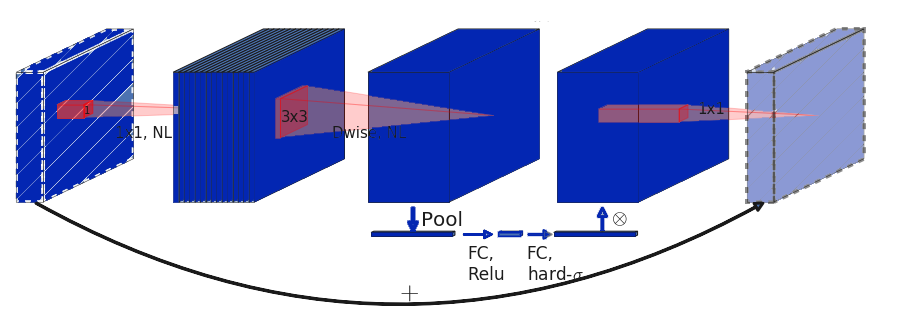}}{}
\makeatother 
\caption{{MobileNetV3 block}}
\label{f-e1b2b17199a1}
\end{figure}

\mbox{}\protect\newline  MobileNetV3 solves the challenges faced by MobileNetV2  by introducing the squeeze and excitation strategy, which gives unequal weights to different input channels as opposed to equal weights from a normal CNN. In contrast to the separate addition of the Squeeze and excitation layer in the 'resnet' block, it applies them in parallel for improving performance and accuracy. Figure~\ref{f-e1b2b17199a1} shows the MobileNetV3 block diagram.

The results section has briefly discussed the comparison results of variants of MobileNet on the food detection dataset leading to the conclusion that MobileNetV3 selected for feature extraction has superior performance on the food detection dataset in contrast to other variants of MobileNet. Following feature extraction selecting a reliable set of features from a high dimensional space of an efficient neural network is vital to improve overall performance. The following section discusses our method for reliable feature selection from high dimensional space to achieve the objective of optimal performance.

\subsection{\textbf{Gradient Explainer for Feature Selection From High Dimensional Feature Vectors}}For reliable feature selection, the proposed method uses SHAP values from GradientExplainer\unskip~\cite{1092609:22126329}. The GradientExplainer combines the ideas from integrated gradients, SHAP, and Smooth gradients into a single expected equation and uses the entire dataset as a background distribution instead of a single reference value after allowing local smoothing for approximating the model as a linear function between each background sample and currently explained input. 

\begin{figure}[!ht]
\centering
\includegraphics[width=5in]{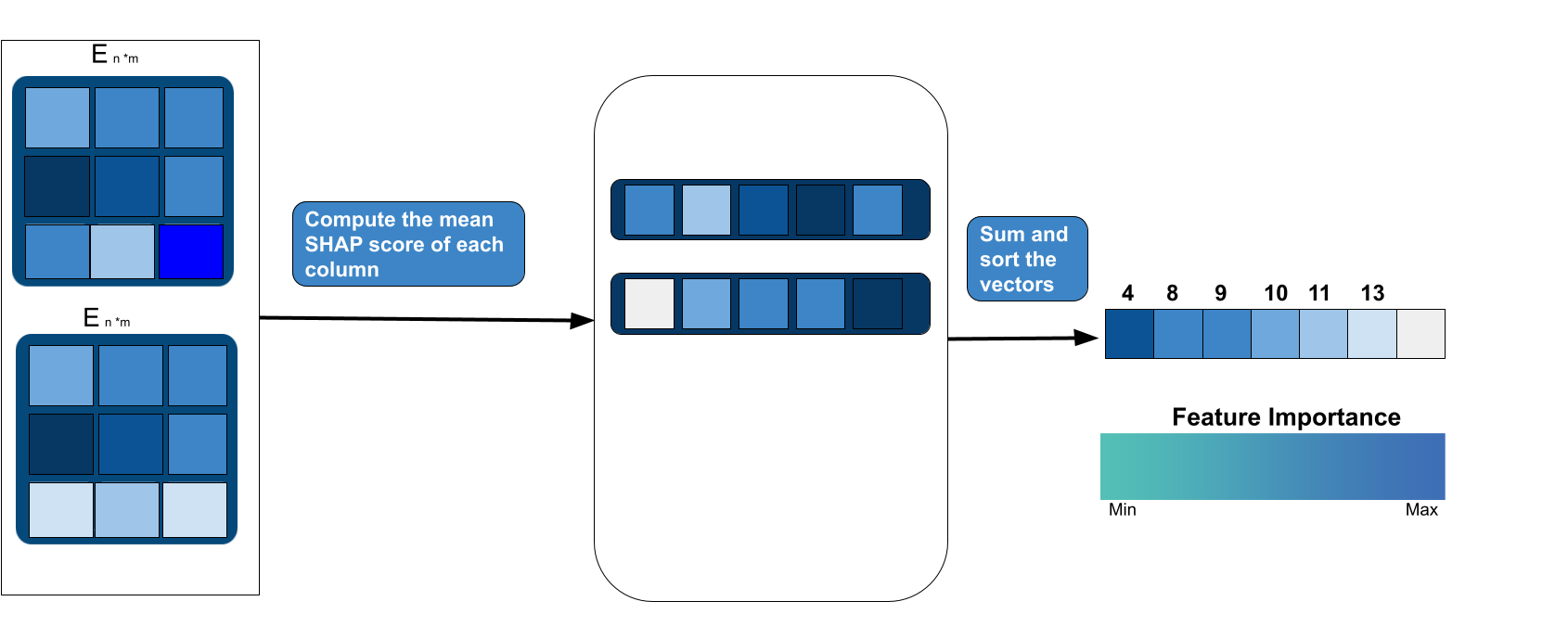}

\caption{Shows the process of feature selection.}
\label{f-c3cb47e6a47e}
\end{figure}

The proposed method firstly uses the GradientExplainer method to compute the SHAP score of features, showing the importance of features on the output layer.  There on,  proposed method has calculated the SHAP score from the pooling layer (3,3,1280) of the MobileNetV3 for training images and then applies average pooling to get a single vector. Equation 1 has aggregated the SHAP scores of the features of each class, while Equation 2 computes average mean SHAP score of each class. \mbox{}\protect\newline Based on the aggregated score, the method ranked the features, determined the indexes of the top 500 attributes with the maximum SHAP score.  Figure~\ref{f-c3cb47e6a47e} showed the process for attribute selection.
\let\saveeqnno\theequation
\let\savefrac\frac
\def\dispfrac{\displaystyle\savefrac}
\begin{eqnarray}
\let\frac\dispfrac
\gdef\theequation{1}
\let\theHequation\theequation
\label{dfg-f0ace8576455}
\begin{array}{@{}l}\varphi^c=\;{\textstyle\frac{\sum_{i=1}^{i=t}\phi^i}t}\end{array}
\end{eqnarray}
\global\let\theequation\saveeqnno
\addtocounter{equation}{-1}\ignorespaces 

\let\saveeqnno\theequation
\let\savefrac\frac
\def\dispfrac{\displaystyle\savefrac}
\begin{eqnarray}
\let\frac\dispfrac
\gdef\theequation{2}
\let\theHequation\theequation
\label{dfg-a1f0f2fafdf9}
\begin{array}{@{}l}A=\;{\textstyle\frac{\sum_{i=1}^{i=n}\varphi^i}n}\end{array}
\end{eqnarray}
\global\let\theequation\saveeqnno
\addtocounter{equation}{-1}\ignorespaces 
During the testing phase, features are selected corresponding to the indexes with the maximum Shap score on the training dataset.

\subsection{\textbf{Parallel Multiclassification for Big Data}}Following efficient feature extraction and reliable feature selection, the final phase in our framework is highly accurate food detection. To achieve the objective, we proposed a novel method for food detection by inheriting the deep representation of the CNN and the approximability of KELMs. Compared to the linear-fully connected classifier, KELM discriminates features better due to its nonlinear function. However, large datasets suffer from the curse of dimensionality. Reduced kernel extreme learning machine (RKELM) reduces the dimensionality of the kernel matrix by randomly selecting the 10 percent of nodes as mapping nodes. However, it is not suitable due to variation in the classification performance. Especially for a performance-critical task like food/non-food classification, detecting the non-food image as food or vice versa affects the usability. Furthermore, RKELM also suffers from the dimensionality problem for a huge dataset. To solve this problem, our introduced MCKELM uses the k-d tree algorithm to divide the data into multiple subsets and then train separate KELM on each subset. A brief discussion of the background and mathematical working of the proposed method for food detection is as follows. Table~\ref{tw-4996acb96f1c} and Table~\ref{tw-7f76d0a28e58} shows the acronyms definition and notation definitions.

\begin{table}[!htbp]
\caption{{Acronyms definition} }
\label{tw-4996acb96f1c}
\def\arraystretch{1}
\ignorespaces 
\centering 
\begin{tabulary}{\linewidth}{LL}
\hline Symbol & Description\\
\hline 
ELM &
  extreme learning machine\\
KELM &
  kernel extreme learning machine \\
k-d tree &
  k dimensional tree \\
MCKELM &
  Multi column kernel extreme learning machine\\
\hline 
\end{tabulary}\par 
\end{table}

\begin{table}[!ht]
\caption{{Notation definition} }
\label{tw-7f76d0a28e58}
\def\arraystretch{1}
\ignorespaces 
\centering 
\begin{tabulary}{\linewidth}{LL}
\hline Symbol & Description\\
\hline 
t &
  Total number of images in each class.\\
c &
  Total number of classes.\\
$ \varphi $&
  SHAP score of all attributes for each class.\\
A &
   SHAP score vector of all attributes.\\
$\eta $ &
   Number of chopping process.                 \\
N &
   Number of resultant subsets.  \\
 C &
  Single chop of the dataset. \\
$D^{i} $ &
  Average density of chopped dataset.\\
O &
  Original dataset density.\\
$ \mathfrak H $ &
  Denotes single subser in MCKLEM.\\
$ x_k $ &
  Training input vector of the kth training data set.\\
$ \widehat x $ &
  input vector\\
$ x^{i} $ &
  ith element of the input vector x\\
$ K $ &
  kernel matrix of KELM\\
$ \beta $ &
  Output Weights \\
Y &
  Output Labels\\
\hline 
\end{tabulary}\par 
\end{table}

\subsubsection{\textbf{Kernel Extreme Learning Machine}}
 KELM is a unified learning model where they removed optimization restrictions of support vector machine (SVM), least-squares support vector machines (LS-SVM), and proximal support vector machines (PSVM)  from the bias term. With lower optimization constraints, the generalization performance is enhanced, while also reducing the complexity. \mbox{}\protect\newline According to Huang et al. \unskip~\cite{1092609:22126334} if hidden layer `h.' is unknown, they can apply Mercer's conditions on an extreme learning machine (ELM) by defining the kernel matrix for ELM by using eq. ~(\ref{dfg-f0c16c21e916}).
\let\saveeqnno\theequation
\let\savefrac\frac
\def\dispfrac{\displaystyle\savefrac}
\begin{eqnarray}
\let\frac\dispfrac
\gdef\theequation{3}
\let\theHequation\theequation
\label{dfg-f0c16c21e916}
\begin{array}{@{}l}K_{ELM}\;=\;HH^{T}\;:\;K_{ELMi,j}\;=\;h(x_i)\;=\;h(x_j)\;=\;K(x_i,\;x_j)\end{array}
\end{eqnarray}
\global\let\theequation\saveeqnno
\addtocounter{equation}{-1}\ignorespaces 
Where,  $\;k(x_i,y_i) $ is the output of the single-layer hidden neural networks.

The output function of KELM is denoted by eq. ~(\ref{dfg-85432fa8dd71})
\let\saveeqnno\theequation
\let\savefrac\frac
\def\dispfrac{\displaystyle\savefrac}
\begin{eqnarray}
\let\frac\dispfrac
\gdef\theequation{4}
\let\theHequation\theequation
\label{dfg-85432fa8dd71}
\begin{array}{@{}l}f(x)\;=\;\begin{bmatrix}\begin{array}{c}k(x,x_1)\\\vdots\end{array}\\k(x,x_n)\end{bmatrix}^{T}\;(CI+K_{KELM})^{-1}T\end{array}
\end{eqnarray}
\global\let\theequation\saveeqnno
\addtocounter{equation}{-1}\ignorespaces 
According to Frenay and Verleysan \unskip~\cite{1092609:22126330} when h(x) is known, KELM can be defined by using eq. ~(\ref{dfg-bbee8eac3df6}).
\let\saveeqnno\theequation
\let\savefrac\frac
\def\dispfrac{\displaystyle\savefrac}
\begin{eqnarray}
\let\frac\dispfrac
\gdef\theequation{5}
\let\theHequation\theequation
\label{dfg-bbee8eac3df6}
\begin{array}{@{}l}k(u,v)\;=\;\lim_{L\rightarrow\infty}\frac1Lh(u).h(v)\end{array}
\end{eqnarray}
\global\let\theequation\saveeqnno
\addtocounter{equation}{-1}\ignorespaces 
There are various kernels from the existing research work that satisfies mercer's condition. This research work has used the radial basis function (RBF) kernel and chi-square kernel.
1) RBF Kernel
\let\saveeqnno\theequation
\let\savefrac\frac
\def\dispfrac{\displaystyle\savefrac}
\begin{eqnarray}
\let\frac\dispfrac
\gdef\theequation{6}
\let\theHequation\theequation
\label{dfg-efe71cbd482e}
\begin{array}{@{}l}k(\mu_i,\sigma _i,\gamma,x)\;=\;exp\left(\frac{\gamma\left\ensuremath{\Vert}x-\mu_i\right\ensuremath{\Vert}^{2}}{\sigma _i^{2}}\right)\end{array}
\end{eqnarray}
\global\let\theequation\saveeqnno
\addtocounter{equation}{-1}\ignorespaces 
2) Chi-square Kernel
\let\saveeqnno\theequation
\let\savefrac\frac
\def\dispfrac{\displaystyle\savefrac}
\begin{eqnarray}
\let\frac\dispfrac
\gdef\theequation{7}
\let\theHequation\theequation
\label{dfg-bb498989f932}
\begin{array}{@{}l}k(\mu_i,\sigma _i,\gamma,x)\;=\;exp\left(-\sigma \;Sum\;\left[\frac{(x\;-\;\mu)^{2}\;}{(x\;+\mu)}\right]\;\right)\end{array}
\end{eqnarray}
\global\let\theequation\saveeqnno
\addtocounter{equation}{-1}\ignorespaces 
The features must be normalized between 0 and 1 when using the chi-square kernel, as $\mu $ must contain values for this kernel. 

\begin{figure}[!ht]
\centering
\includegraphics[width=5in]{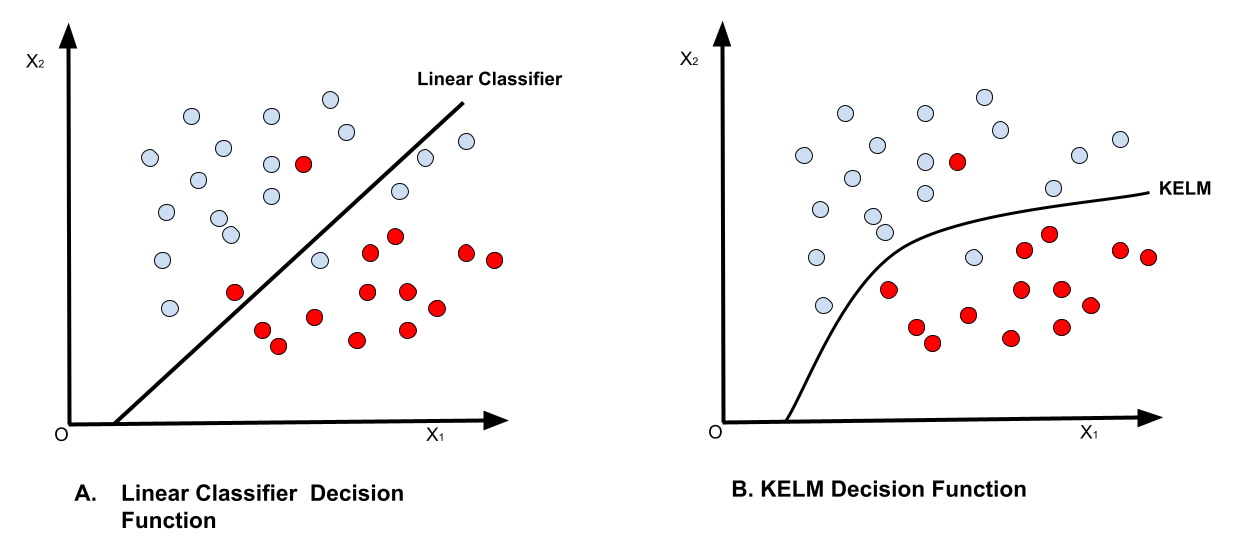}

\caption{Visual illustration of the decision boundary of linear classifier and KELM.}
\label{f-e25b3367604a}
\end{figure}

In the above kernel adjustable parameter,  $\sigma  $ significantly affects the classifier performance and should be adjusted based on the dataset. In contrast to the ELM algorithm,  knowing the feature mapping in the hidden layer is not required, and there is no need to select the number of hidden neurons. Compared to the linear-fully connected classifier, KELM has a superior ability for discriminating feature categories. Figure~\ref{f-e25b3367604a} shows the visual comparison between KELM and linear fully connected layer.

\subsubsection{k-d tree} k-d tree is an important data structure used to store a finite set of points in a k-dimensional space. It is a primarily multi-dimensional binary tree, proposed by Jon Louis Bentley in 1975, which was originally introduced to solve multiple problems emerging in geometric databases\unskip~\cite{1092609:22126336}. Additionally, the k-d tree is well-known for curbing the construction of subsets with zero data and ensuring uniform data distribution, hence enhancing the speed of learning in neural networks \unskip~\cite{1092609:22126346}

For food detection tasks k-d tree algorithm produces subsets of training data to avoid excessive computations and time delays resulting from a high number of hidden nodes in the single kernel matrix. \mbox{}\protect\newline  \mbox{}\protect\newline Since each non-leaf node in the k-d tree creates a binary-splitting hyperplane. it distributes the points into left and right sub-trees, with each leaf node representing a k-dimensional point. Consequently, it reduces network training and testing time due to fewer hidden nodes in each subset. Each subset act as independent training data set for each KELM. In the k-d tree, subsets possess the same density as the original dataset and contain enough instances for training purposes. Algorithm 1 explains the steps for chopping the dataset into subsets and the training process of the KELM algorithm on these subsets.

% \bgroup
% \fixFloatSize{images/kdtreediv.png}
% \begin{figure}[!htbp]
% \centering \makeatletter\IfFileExists{images/kdtreediv.png}{\includegraphics[width=3.1in]{images/kdtreediv.png}}{}
% \makeatother 
% \caption{{3-dimensional visualization of the k-d tree shows the splitting process. The first split (red) cuts the root dataset into two subsets. Each of these subsets is further split (green) into two subsets. As we cannot divide it further, the final eight subsets are called leafs (final subsets).}}
% \label{f-c8898c2ad904}
% \end{figure}
% \egroup
\begin{figure}[!ht]
\centering
\includegraphics[width=3.1in,height=2.7in]{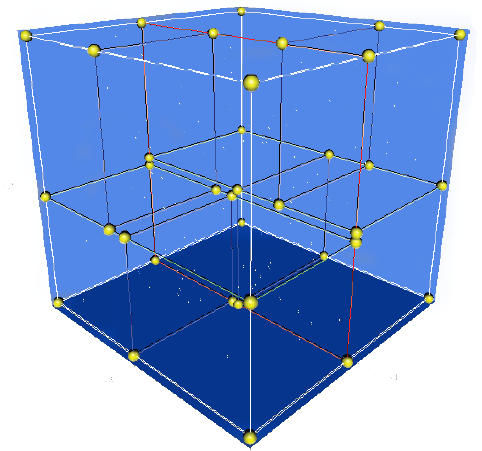}

\caption{3-dimensional visualization of the k-d tree shows the splitting process. The first split (red) cuts the root dataset into two subsets. Each of these subsets is further split (green) into two subsets. As we cannot divide it further, the final eight subsets are called leafs (final subsets).}
\label{f-c8898c2ad904}
\end{figure}

\subsubsection{\textbf{Multicolumn Kernel Extreme Learning Machine}} 
\begin{figure}[!ht]
\centering
\includegraphics[width=5.0in]{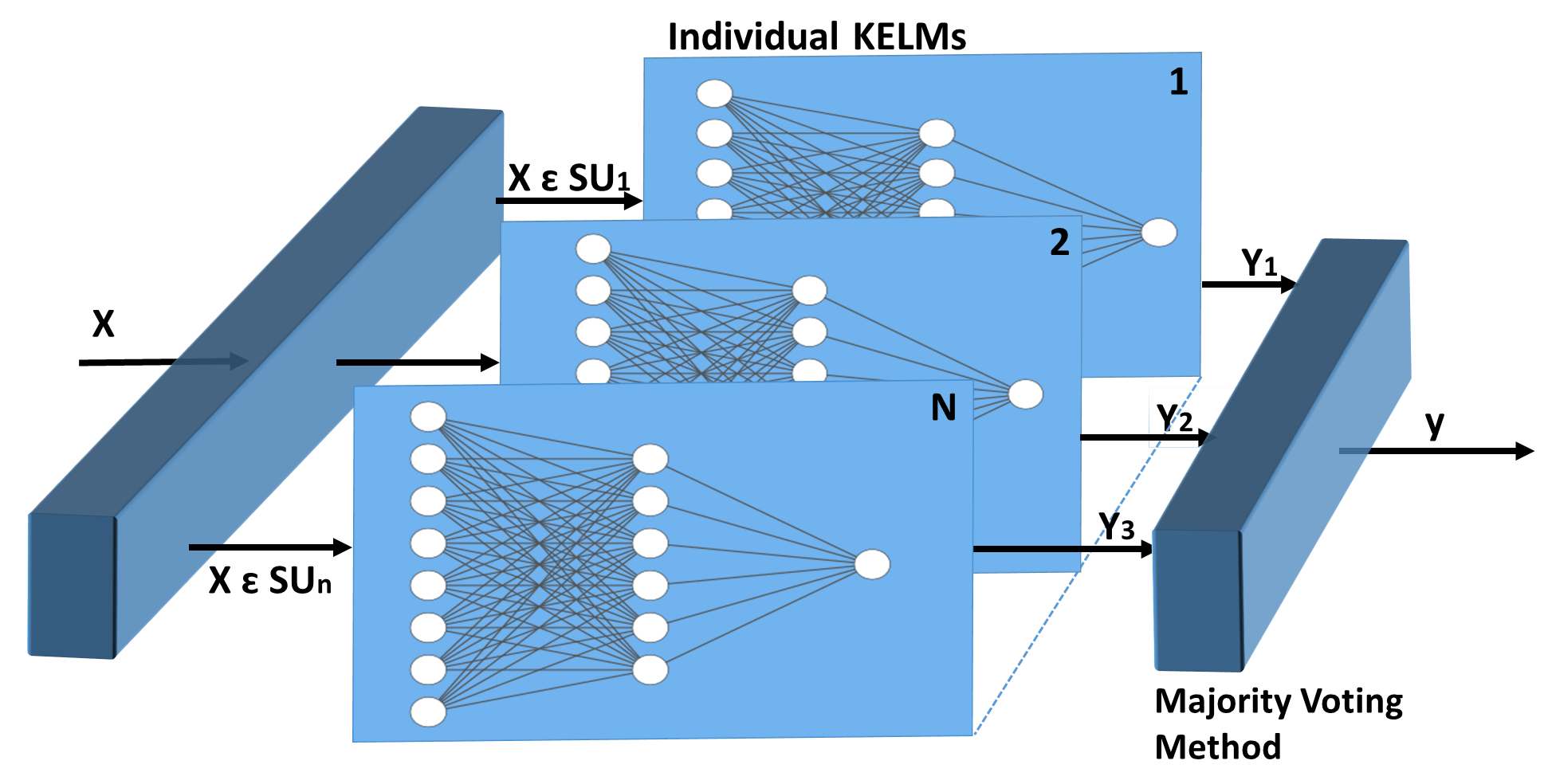}

\caption{Multicolumn Kernel Extreme Learning Machine}
\label{f-9658c0bcac4a}
\end{figure}

\begin{figure}[!ht]
\centering
\includegraphics[width=3.7in]{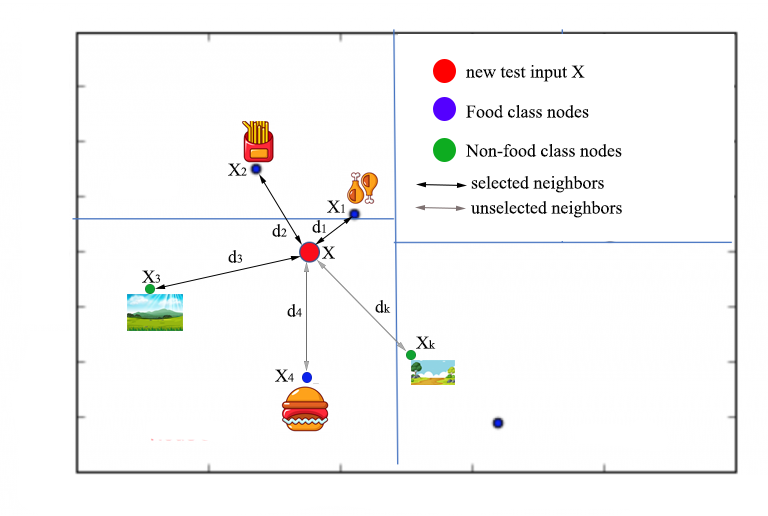}

\caption{Selection of k subsets and k KELM based on the KNN algorithm}
\label{f-3ac00d890be8}
\end{figure}

\mbox{}\protect\newline The proposed multicolumn kernel extreme learning machine network (MCKELM) method reduces the complexity of inner matrix computations during the training and testing process in KELM.  It uses parallel structures and requires fewer hidden units per KELM and fewer instances of datasets to train each KELM. MCKELM operates by dividing the large dataset into smaller subsets using the k-d tree algorithm and considers the resulting 'N' subsets as individual datasets required to train 'N' separate KELM. Consequently, the individual KELMs are then stacked in a parallel manner as shown in Figure~\ref{f-9658c0bcac4a} and it only selects top k subsets during the testing phase of MCKELM. \mbox{}\protect\newline As it trains KELM  on their dataset in isolation from other KELMs, MCKELM is considered a manageable and highly developed parallel structure instead of traditional KELM, which suffers from the drawback of computing-intensive kernel inner product calculations. Thus, MCKELM is an ideal approach for transitioning from a single large structure to a parallel mechanism. \mbox{}\protect\newline During testing, when the model processes a feature vector of the input image, the input subset selector using the k-d tree forwards it to the appropriate KELMs.  Following this, the selected KELMs give their output using the majority voting method to compute the final results. We described the three stages as follows. Algorithm 2 shows the step of the testing process. \mbox{}\protect\newline \textbf{1. Individual subset selector:} \mbox{}\protect\newline Whenever a new instance comes, the input subset selector selects K KELMs based on the k-d tree algorithm.  The k-d tree algorithm determines the euclidean distance between the test vectors and all the nodes in subsets by using Equation~(\ref{dfg-9dae8be4d794}). $d_k $denotes the kth minimum euclidean distance between the test input vector  $\widehat x $and kth training node vector $x_k $; $\widehat x^{i} $ is the ith value of the input vector $\widehat x $ and  $x_n^{i} $ is the ith value of the nth training vector $x_n $. It then selects k nearest vectors with minimum euclidean distance d as shown in Figure~\ref{f-3ac00d890be8}. The selected vectors belong to K subsets, and it selects only KELMs trained on these subsets during the food detection.
\let\saveeqnno\theequation
\let\savefrac\frac
\def\dispfrac{\displaystyle\savefrac}
\begin{eqnarray}
\let\frac\dispfrac
\gdef\theequation{8}
\let\theHequation\theequation
\label{dfg-9dae8be4d794}
\begin{array}{@{}l}d_k=\underset{1...k}{min}(\left\ensuremath{\Vert}\widehat x\;-\;x_n\right\ensuremath{\Vert})\;=\underset{1...k}{min}\left(\sqrt{\sum_{i=1}^{I}}(\widehat x^{i}-x_n^{i})^{2}\right)\end{array}
\end{eqnarray}
\global\let\theequation\saveeqnno
\addtocounter{equation}{-1}\ignorespaces 
\mbox{}\protect\newline \textbf{2. Individual KELMs:} \mbox{}\protect\newline Every `k' neighbor belongs to the subset of the entire dataset required to train separate KELMs, and only selected nearest KELMs work in a parallel manner to generate `n' results. \mbox{}\protect\newline \textbf{3. Output combiner:} \mbox{}\protect\newline It only selects the output from K nearest KELMs. The remaining KELMs do not contribute to the output result. MCKELM computes the final output result by using the simple majority method. \mbox{}\protect\newline MCKELM method is also significantly advantageous in reducing testing time due to parallelized structure in contrast to large connected KELM as smaller subsets essentially avoid generalization problems and instead specify it by informing the MCKELM about the regional experience. Moreover, MCKELM also improves the training and testing times while having superior performance, compared to the softmax classifier. In the results section, we briefly discuss the performance of the classifier.

\newpage
\begin{lstlisting}[style=listing_style,caption={Dividing food/non-food dataset into N subsets and training of KELMs}]
[@\textbf{Input:}@]
[@$\eta $@] Number of chopping process. 
Deep Features of food/non-food training dataset.  
Kernel Extreme Learning Machine model.
[@\textbf{Output:
}@]
[@$2^\eta\;\ast\;KELM $@] trained models of individual KELM         
[@\textbf{Algorithm:}@]
[@$N\;=\;2^\eta $@] Number of subsets and KELM model.
Compute original density of dataset  O.
for [@$ \mathbb{C} $@] = 1 to  [@$\eta $@]
   for subset= 1 to [@$(2^{\mathbb{C}}-1) $@]
     Find the median of each feature
     Compute average density [@$D^i $@]of chopped dataset
     Choose feature with minimum  [@$\left|D^i-O\right| $@] 
     Divide along the median in to two subsets.
   end for
end for
for subset = 1 to [@$\eta $@]
   Train KELM on the subset
end for

\end{lstlisting}

\newpage
\begin{lstlisting}[style=listing_style,caption={Testing process }]
[@\textbf{Input}@]: 
Test Input
[@\textbf{Output}@]: 
y, classification results 
[@\textbf{Algorithm:}@]
Load [@$2^\eta $@] KELM model, where [@$\eta $@] is the total number of chops
for n  = 1 to t
    find the k nearest training nodes to the [@$\widehat x $@] 
    for each subset belonging to k nearest nodes
        Compute the output [@$y_k $@] using KELM trained on the subset    
     end for
end for
Majority voting method for final result [@$y $@]
\end{lstlisting}

\section{Experiments and Results}
This section discusses the food/non-food dataset gathered for this study to evaluate the food detection framework, implementation of the framework, and detailed results of our analysis of each phase. In the first experiment, we determined the best feature extractor from the three variants of the MobileNet for food detection. The second experiment shows that the feature selector method based on Gradient Explainer reduces the training time. The third experiment analyzes the classification performance of MCKELM in contrast to other state-of-the-art methods. Finally, we presented interpretations of the features by using the Shapley-value-based explanations.

\subsection{Food/non-food Dataset} 
\begin{figure}[!ht]
\centering
\includegraphics[width=5.2in]{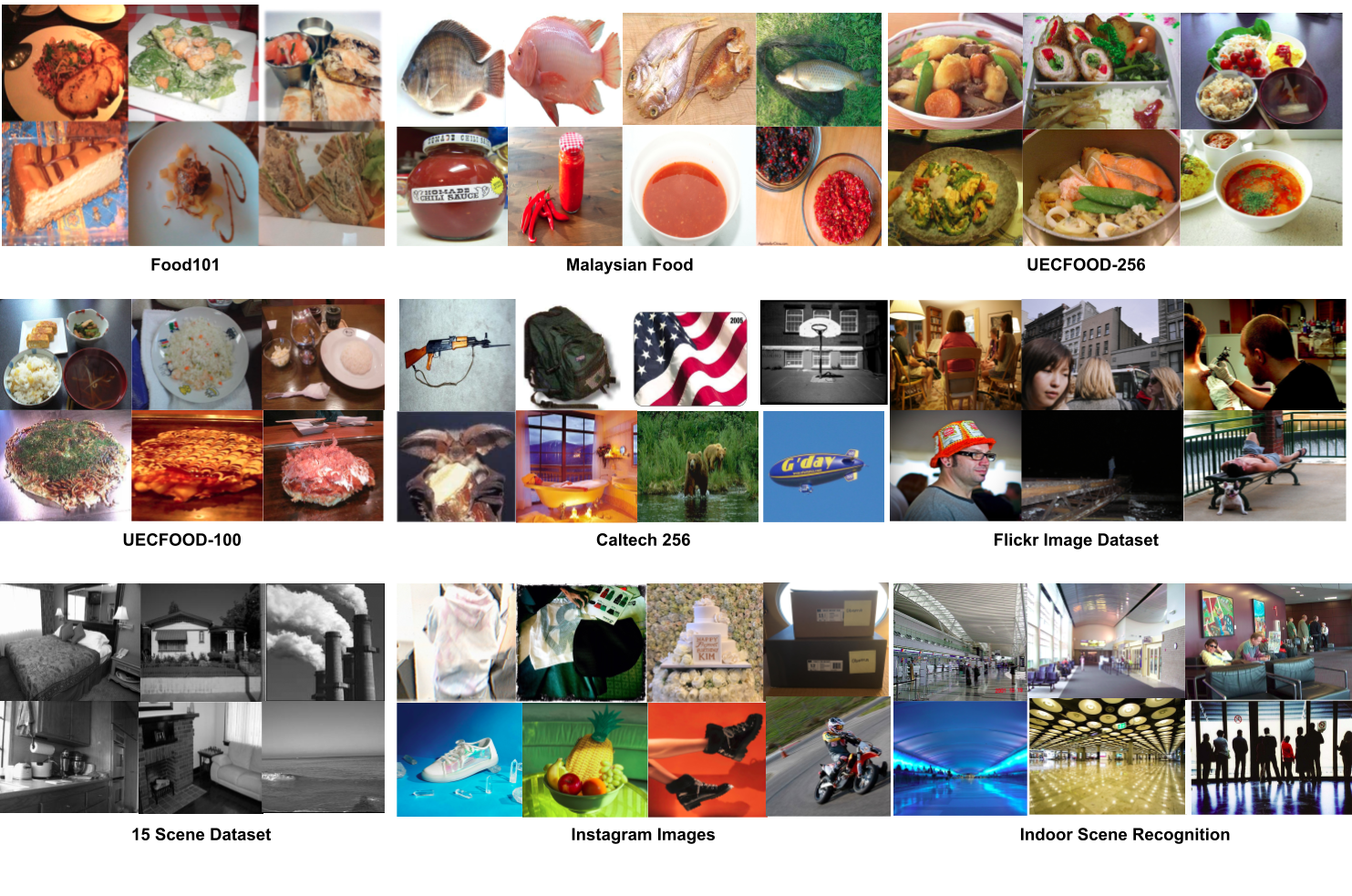}

\caption{Sample images from various dataset used to prepare food detection dataset.}
\label{f-d1b82270d7bccc}
\end{figure}

\begin{figure}[!ht]
\centering
\includegraphics[width=5.0in]{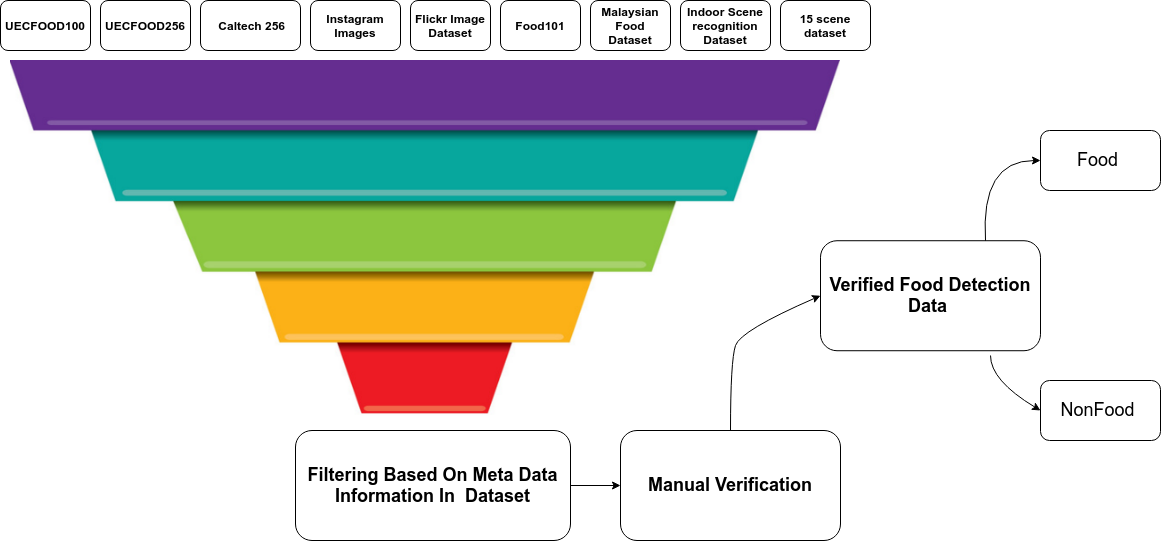}

\caption{Food detection dataset preparation process.}
\label{fooddetectiondatasetpreparation}
\end{figure}

\begin{figure}[!ht]
\centering
\includegraphics[width=5.0in]{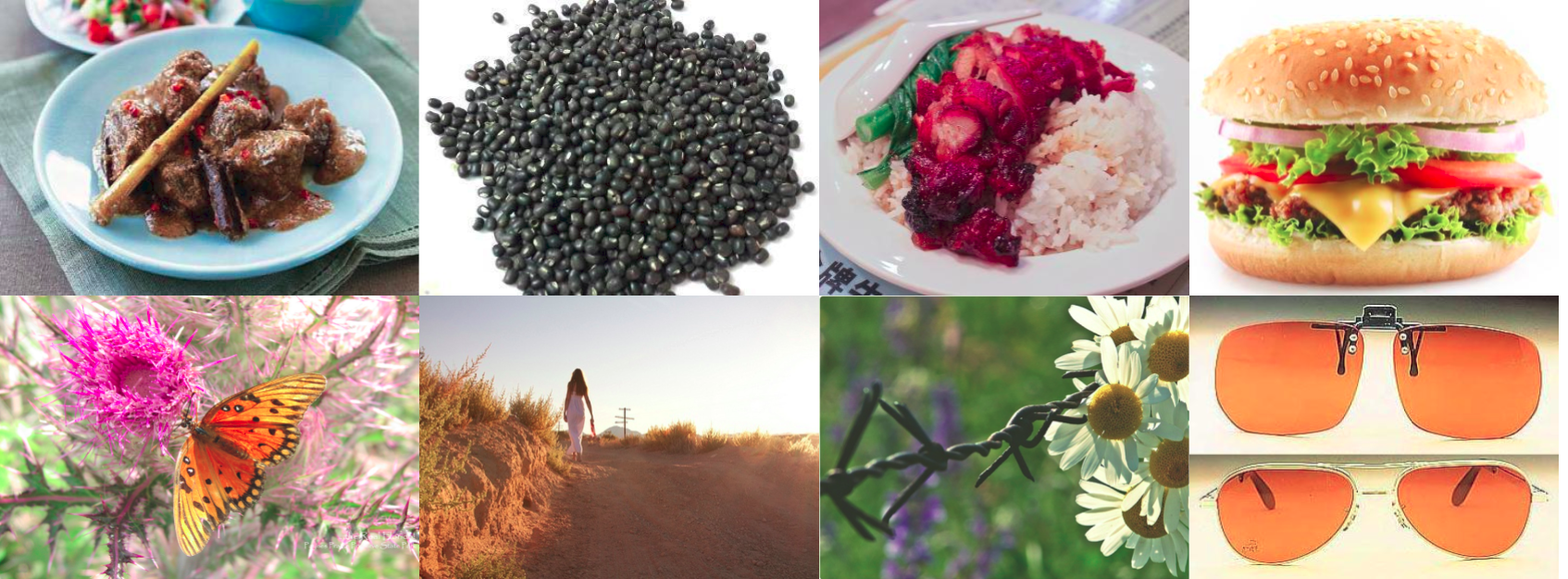}

\caption{Sample images from food detection dataset.}
\label{f-d1b82270d7bc}
\end{figure}

\mbox{}\protect\newline 
We prepared a large-scale food/non-food dataset by using the existing image datasets of UECFOOD100 \unskip~\cite{1092609:22126351}, UECFOOD256\unskip~\cite{1092609:22126352}, Caltech 256 \unskip~\cite{1092609:22126345}, Instagram Images, Flickr Image Dataset \unskip~\cite{1092609:22126521} , Food101\unskip~\cite{1092609:22126343}, Malaysian Food Dataset, Indoor Scene recognition Dataset \unskip~\cite{1092609:22126342} , 15 scene dataset. Inclusion of datsets of various types and from various sources has introduced the diversity in the food detection dataset. As these dataset includes food items of various cultures and images from popular social media platforms. Figure \ref{f-d1b82270d7bccc} shows the sample images from included datasets.
To prepare a dataset for food detection, we define the process which initially filtered the images based on their metadata information. There on, an expert has manually verified food images of the dataset with unclear metadata information to improve the authenticity of the food detection dataset. Figure \ref{fooddetectiondatasetpreparation} diagrammatically illustrates the process of preparation of the food detection dataset.
After preparation, the food detection dataset consists of two classes: food class and non-food class. It has 222430 images for the training and 55096 images for testing, and we open-sourced the dataset for research purposes\unskip~\cite{1092609:22126337}. To the best of our knowledge, it is the largest publically available food/non-food dataset, hence making it suitable for verifying the effectiveness of our proposed multicolumn approach in solving the 'curse of dimensionality problem' faced by KELM. Figure~\ref{f-d1b82270d7bc} shows the sample images from our dataset.

\begin{table}[!htbp]
\caption{{Attributes of food/non-food dataset} }
\label{tw-5614d5e4ca42}
\def\arraystretch{1}
\ignorespaces 
\centering 
\begin{tabulary}{\linewidth}{LLLL}
\hline Dataset & Total class & Total instance & Train/Test instance\\
\hline 
food/non-food &
  2 &
  277,526 &
  222,430/55,096\\
\hline 
\end{tabulary}\par 
\end{table}

\subsection{Implementation}We experimented on the Google Colab (12 GB GPU) and 12 GB random access memory (RAM) to determine the best feature extractor from three variants of the MobileNet. Then for feature extraction and feature selection, we have also used Google Colab. Finally, we trained our classifier using an AMAZON EC2 instance with 64 GB of memory. The trained model is accessible through a web service implemented in python Django.

\subsection{Efficient Feature Extraction Using MobileNet}
We compared MobileNetV1, MobileNetV2, and MobileNetV3 for feature extraction after selecting the network for our framework with the highest classification performance by finetuning all the models for ten epochs at a learning rate of 0.01 and a momentum value of 0.9. The kernel regularizer parameter l2 is set to 0.0005 as shown in Table~\ref{tw-6ca3e76729aa} .  We used stochastic gradient descent (SGD) optimizer, categorical cross-entropy, and softmax activation function. Table~\ref{tw-48c1a376c289} shows the experimental configurations and online data augmentation used for increasing the generalization while Table~\ref{tw-0c8afe99936e} shows the comparison for classification performance. MobileNetV3 outperforms its predecessors in terms of accuracy, and it is faster in contrast to the previous architectures of MobileNet. For this reason, we employed MobileNetV3 in our final framework.

\begin{table}[!ht]
\caption{{Parameter settings} }
\label{tw-6ca3e76729aa}
\def\arraystretch{1}
\ignorespaces 
\centering 
\begin{tabulary}{\linewidth}{LL}
\hline Parameter & Value\\
\hline 
Epochs  &
  10\\
Learning Rate &
  0.01\\
Momentum &
  0.9\\
L2 &
  0.0005\\
Optimizer  &
  SGD\\
\hline 
\end{tabulary}\par 
\end{table}

\begin{table}[!ht]
\caption{{Experimental settings of data augmentation} }
\label{tw-48c1a376c289}
\def\arraystretch{1}
\ignorespaces 
\centering 
\begin{tabulary}{\linewidth}{LL}
\hline Data-Augmentation Type & Value\\
\hline 
Wide Shift &
  0.2\\
Height Shift &
  0.2\\
Horizontal Shift &
  Randomly flips horizontally\\
Zoom Range &
  0.8 to 1\\
Channel Shift Range &
  30\\
Full Mode &
  Reflect\\
\hline 
\end{tabulary}\par 
\end{table}

\begin{table}[!ht]
\caption{{Comparison of classification accuracy on linear fully connected classifier} }
\label{tw-0c8afe99936e}
\def\arraystretch{1}
\ignorespaces 
\centering 
\begin{tabulary}{\linewidth}{LLLL}
\hline Datasets & MobileNetV1 & MobileNetV2 & MobileNetV3\\
\hline 
food/non-food &
  98.33 &
  98.56 &
  \textbf{98.73}\\
\hline 
\end{tabulary}\par 
\end{table}

\subsection{Gradient Explainer for Image Feature Selection From High Dimensional Data}
Feature vectors extracted from the CNN model have a very high dimension.   \mbox{}\protect\newline This work employed the strategy which uses SHAP values from GradientExplainer to select the best top 500 features. The comparison of the training time, and testing time in Table~\ref{tw-3a435ea43fd3} shows that the feature selection reduces the computational complexity while having competitive classification performance. The reduction in training time is significant when the total number of subsets is less in MCKELM.  However, this difference decreases when subsets are more due to reduced complexity of inner matrix computations, as experimental analysis shows that it reduces testing time by 77.14\% for 16 subsets and 52\% for 32 subsets.

\begin{table*}[!ht]
\caption{{Comparison of Training/Testing Time by using SHAP feature selection} }
\label{tw-3a435ea43fd3}
\def\arraystretch{1}
\ignorespaces 
\centering 
\begin{tabulary}{\linewidth}{p{\dimexpr.20\linewidth-2\tabcolsep}p{\dimexpr.20\linewidth-2\tabcolsep}p{\dimexpr.20\linewidth-2\tabcolsep}p{\dimexpr.20\linewidth-2\tabcolsep}p{\dimexpr.20\linewidth-2\tabcolsep}}
\hline  & \multicolumn{2}{p{\dimexpr(.40\linewidth-2\tabcolsep)}}{Training Time} & \multicolumn{2}{p{\dimexpr(.40\linewidth-2\tabcolsep)}}{Testing Time}\\
Model & All Features & SHAP & All Features & SHAP\\
\hline 
MCKELM \mbox{}\protect\newline (16 Subsets) &
  660 s &
  \textbf{343 s} &
  0.35 s &
  \textbf{0.08 s}\\
MCKELM \mbox{}\protect\newline (32 Subsets) &
  302 s &
  306 s &
  0.23 s &
  \textbf{0.11 s}\\
\hline 
\end{tabulary}\par 
\end{table*}

\subsection{Visual Interpretation of Features From CNN}
\begin{figure}[!ht]
\centering
\includegraphics[width=5.0in]{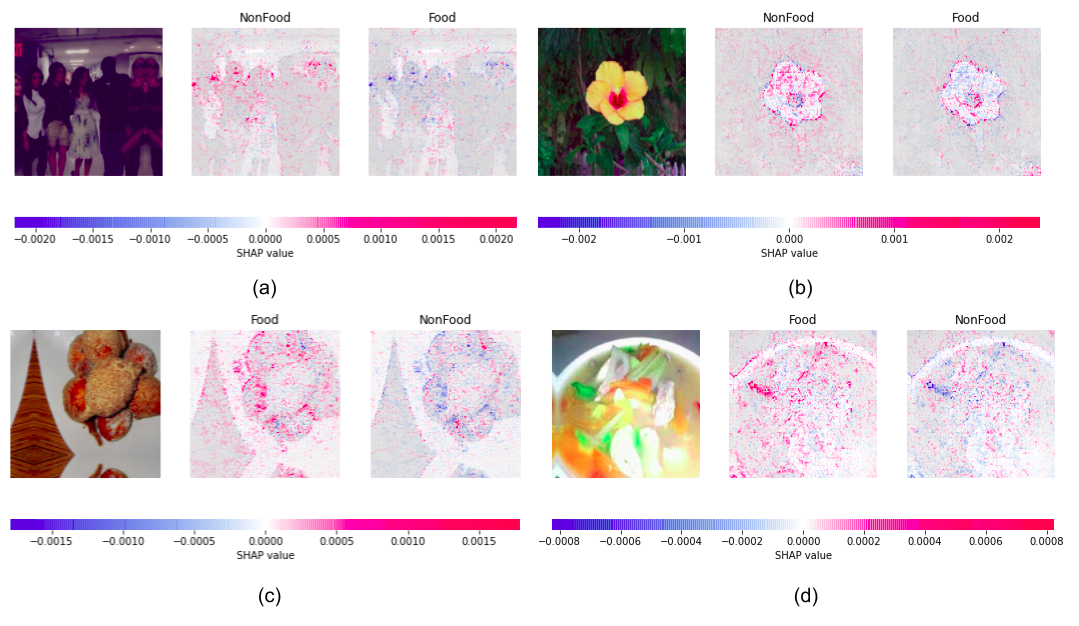}

\caption{The red pixels provide the features that lean the classifier towards positive classification, while the blue pixels provide features that tend the classifier towards the negative class.}
\label{f-2726d0d69fbd}
\end{figure}

We employed 'GradientExplainer' to visualize the pixels and their contribution to the output, as these pixels provide features that cause the classifier to lean towards a particular class. Our fast approximation algorithm computes the game theory-based SHAP values\unskip~\cite{1092609:22126329} which considers multiple background samples instead of a single baseline. In our experiment, we set the number of background samples to 500 after randomly selecting these samples from the training dataset of both classes. We set the ranked output parameter to 2. Figure~\ref{f-2726d0d69fbd} shows the visualizations generated by GradientExplainer method.  The red pixels provide the features that make the classifier lean toward the positive classification, and the blue pixels provide features that make the classifier tends toward the negative classification. These visualizations indicate regions in the image, which tend to bend the decision, hence increasing our confidence in framework predictions by ensuring that the classifier is making decisions based on the features from the desired regions of the photos.

% \bgroup
% \fixFloatSize{images/explainable_ai.png}
% \begin{figure*}[!ht]
% \centering \makeatletter\IfFileExists{images/explainable_ai.png}{\includegraphics{images/explainable_ai.png}}{}
% \makeatother 
% \caption{{The red pixels provide the features that lean the classifier towards positive classification, while the blue pixels provide features that tend the classifier towards the negative class.}}
% \label{f-2726d0d69fbd}
% \end{figure*}
% \egroup
\subsection{Classification Performance on Big Data}This section compares the experimental results of the proposed MCKELM with other classifiers. Section 4.5.1 presents the measures used for analysis, and Section 4.5.2 presents the results of the MCKELM in contrast with other methods.

\subsubsection{Measures}We used the following metrics to measure classification performance.

\paragraph{Accuracy }It measures the ability of the classifier to predict the class accurately. Equation~(\ref{dfg-0ac076b68c3a}) denotes the accuracy of the network.
\let\saveeqnno\theequation
\let\savefrac\frac
\def\dispfrac{\displaystyle\savefrac}
\begin{eqnarray}
\let\frac\dispfrac
\gdef\theequation{8}
\let\theHequation\theequation
\label{dfg-0ac076b68c3a}
\begin{array}{@{}l}Accuracy\;=\frac{\;(TP+FN)}{(TP+FP+FN+TN)}\times100\;\end{array}
\end{eqnarray}
\global\let\theequation\saveeqnno
\addtocounter{equation}{-1}\ignorespaces

\paragraph{F1 Score }It is interpreted as a weighted average of precision and recall as both recall and precision contributed equally to the weighted average.  Equation~(\ref{dfg-7e31ba67dab2}) denotes the F1 Score.
\let\saveeqnno\theequation
\let\savefrac\frac
\def\dispfrac{\displaystyle\savefrac}
\begin{eqnarray}
\let\frac\dispfrac
\gdef\theequation{9}
\let\theHequation\theequation
\label{dfg-7e31ba67dab2}
\begin{array}{@{}l}F1\;Score\;=\;\frac{(2\ast(Precision\ast Recall))}{Precision+Recall}\;\end{array}
\end{eqnarray}
\global\let\theequation\saveeqnno
\addtocounter{equation}{-1}\ignorespaces

\paragraph{Recall }It is the network's ability to find all positive instances. Equation~(\ref{dfg-ab037af9d4f9}) denotes the Recall score.
\let\saveeqnno\theequation
\let\savefrac\frac
\def\dispfrac{\displaystyle\savefrac}
\begin{eqnarray}
\let\frac\dispfrac
\gdef\theequation{10}
\let\theHequation\theequation
\label{dfg-ab037af9d4f9}
\begin{array}{@{}l}Recall\;=\;\frac{TP}{(TP+FN)}\end{array}
\end{eqnarray}
\global\let\theequation\saveeqnno
\addtocounter{equation}{-1}\ignorespaces

\paragraph{Precision }It measures the ability of the network to not label negative samples as positive. Equation~(\ref{dfg-910740dcbd36}) computes the precision of the model.
\let\saveeqnno\theequation
\let\savefrac\frac
\def\dispfrac{\displaystyle\savefrac}
\begin{eqnarray}
\let\frac\dispfrac
\gdef\theequation{11}
\let\theHequation\theequation
\label{dfg-910740dcbd36}
\begin{array}{@{}l}Precision\;\;=\frac{TP}{(TP\;+FP)}\;\;\end{array}
\end{eqnarray}
\global\let\theequation\saveeqnno
\addtocounter{equation}{-1}\ignorespaces

\subsubsection{Results}As food detection is a binary problem, we assumed a batch learning fixed-class scenario for training MCKELM using processed features. In our experiments kernel parameter is set to 1 for MCKELM, and the value of the k-d tree's leaf size in MCKELM is set to 13000 for 32 subsets and 25000 for 16 subsets. In K-nearest neighbor (KNN), the number of the nearest neighbor k value is 1. For RKELM, the kernel matrix randomly selects 10 percent of nodes as mapping nodes. For ELM, the number of hidden neurons value is determined by using the following heuristic.

\begin{figure}[!htbp]
\centering
\includegraphics[width=5.4in]{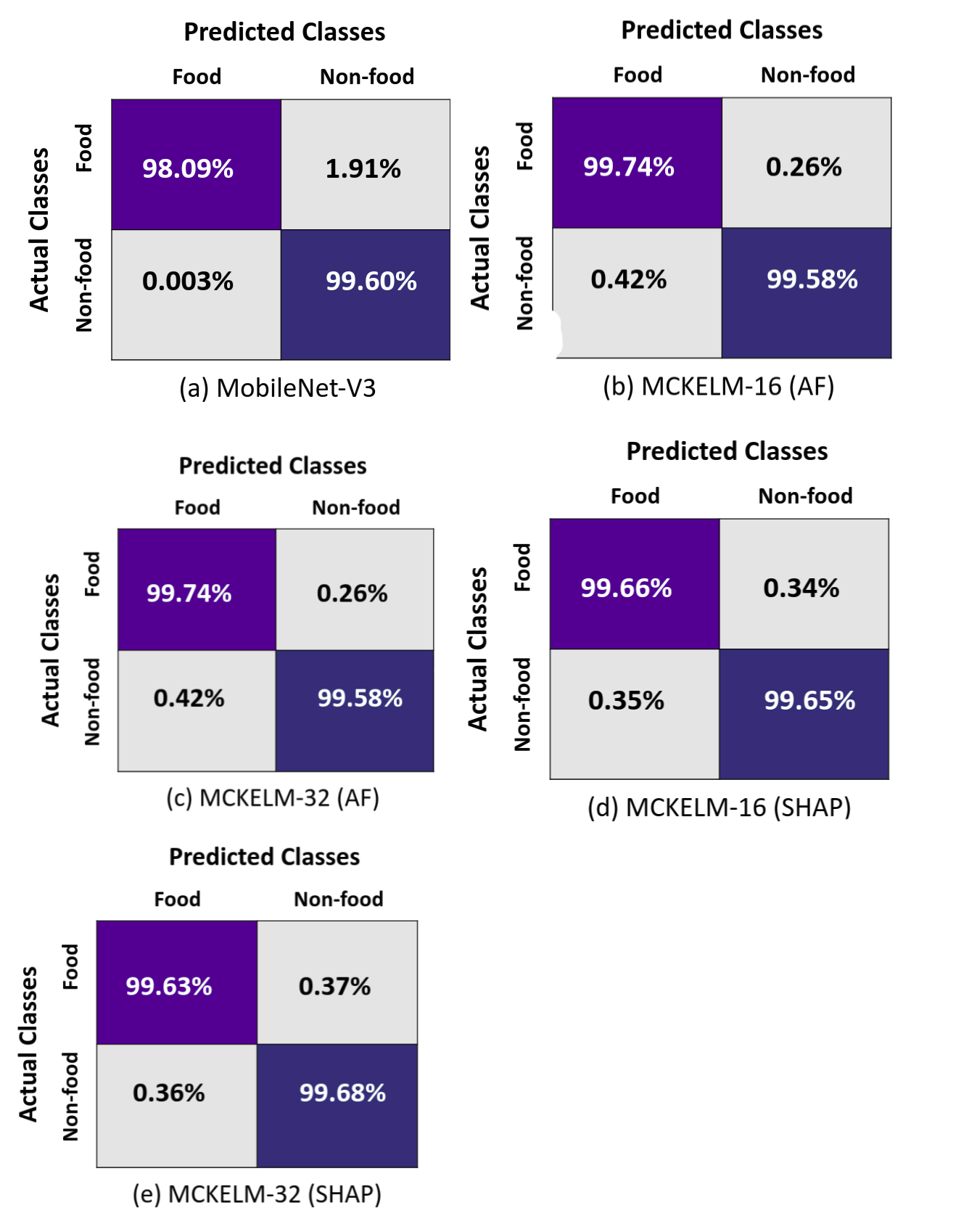}

\caption{Confusion matrix of linear fully connected classifier in MobileNetV3 and MCKELM at 16 subsers and 32 subsers for food/non-food classification.}
\label{f-593489c3aa3c}
\end{figure}

\begin{figure}[!ht]
\centering
\includegraphics[width=5.4in]{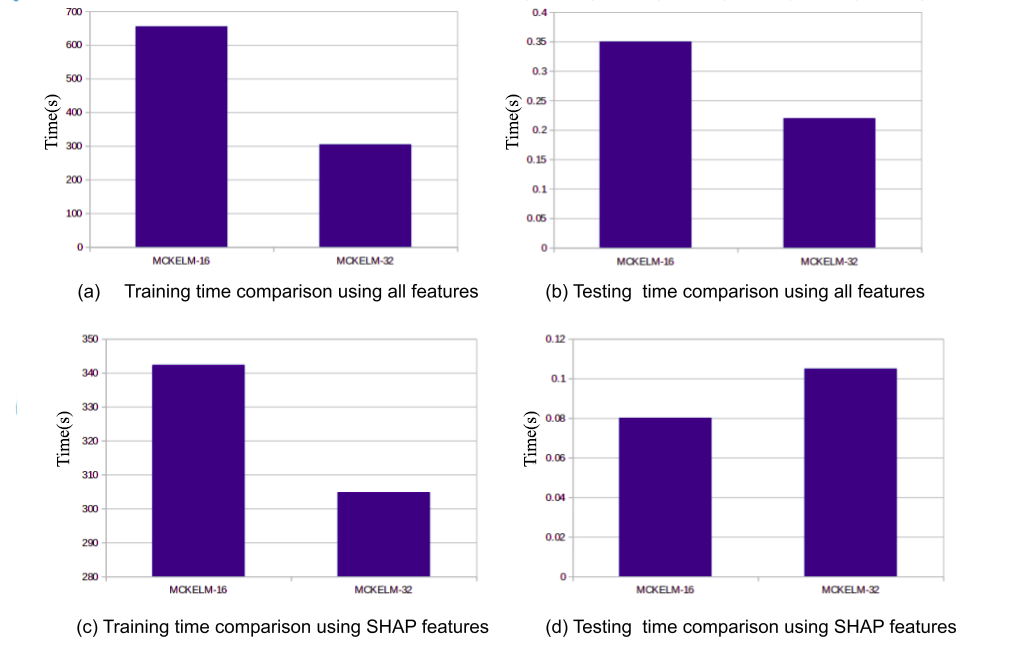}

\caption{Comparison of training and testing time at different subsers of MCKELM}
\label{f-1151c58b60d6}
\end{figure}

 We cannot evaluate the KELM algorithm for such a large dataset due to the curse of dimensionality as it requires more than 384 GB of ram for processing the feature matrix of food/non-food datasets. However, our proposed MCKELM resolves the dimensionality problem faced by KELM. Table 12 compares the classification results of deep learning models and MCKELM. We reported the results of MCKELM for 16 subsets and 32 subsets. The experimental analysis shows that novel MCKELM for classification improves performance compared to MobileNetV3 by 0.93\%. It is worth noting that the dataset gathered by us for food/non-food recognition is large, and the improvement of the performance by MCKELM is significant as it correctly classifies \textbf{513} more images compared to MobileNetV3. The comparison with the Extreme learning machine shows that ELM misclassifies \textbf{66} instances more than MCKELM. Similarly, reduced kernel extreme learning machine has misclassified \textbf{105} instances, k nearest neighbor has misclassified, \textbf{54} instances, and Naive Bayesian classifier misclassified \textbf{109} more instances compared to proposed MCKLEM. Figure \ref{f-593489c3aa3c} presents the confusion matrix of MobileNetV3 and MCKELM at different subsets that further elaborates the improvement by MCKELM.  \mbox{}\protect\newline Figure \ref{f-1151c58b60d6}  shows the comparison of the training and testing time at various subsets of MCKELM. For MCKELM increasing the number of subsets reduces training time and testing time for our case.

% \bgroup
% \fixFloatSize{images/areaundercurve.png}
% \begin{figure}[!htbp]
% \centering \makeatletter\IfFileExists{images/areaundercurve.png}{\includegraphics[width=3.5in]{images/areaundercurve.png}}{}
% \makeatother 
% \caption{{Area under curve for MobileNetV1, MobileNetV2 and MobileNetV3}}
% \label{f-f955f734ff94}
% \end{figure}
% \egroup

\begin{table*}[!htbp]
\caption{{Comparison of food/non-food Classification.} }
\label{tw-8deb64abafa0}
\def\arraystretch{1}
\ignorespaces 
\centering 
\begin{tabulary}{\linewidth}{p{\dimexpr.298\linewidth-1\tabcolsep}p{\dimexpr.21\linewidth-2\tabcolsep}p{\dimexpr.175\linewidth-2\tabcolsep}p{\dimexpr.2117\linewidth-2\tabcolsep}p{\dimexpr.195\linewidth-2\tabcolsep}}
\hline Model & Accuracy & Precision & Recall & F1 Score\\
\hline 
MobileNetV1 &
  98.33 &
  96.76 &
  99.39 &
  98.05\\
MobileNetV2 &
  98.56 &
  97.51 &
  99.14 &
  98.32\\
MobileNetV3 &
  98.73 &
  97.45 &
  99.60 &
  98.52\\
MobileNetV3 (Ensemble) &
  99.12 &
  99.02 &
  99.18 &
  99.10\\  
Naive Baysian (AF) &
  99.00 &
  99.00 &
  99.00 &
  99.00\\
Naive Baysian (SHAP) &
  99.46 &
  99.39 &
  99.49 &
  99.44\\
KNN (AF) &
  99.56 &
  99.56 &
  99.55 &
  99.55\\
KNN (SHAP) &
  99.58 &
  99.56 &
  99.57 &
  99.57\\
ELM (AF) &
  99.58 &
  99.57 &
  99.58 &
  99.58\\
ELM (SHAP) &
  99.54  &
  99.53 &
  99.54 &
  99.55\\
RKELM (AF) &
  98.98 &
  98.89 &
  99.04 &
  98.96\\
RKELM (SHAP) &
  99.47 &
  99.45 &
  99.47 &
  99.46\\
MCKELM (RBF)  \mbox{}\protect\newline (16-subsets) (AF) &
  \textbf{99.66} &
  \textbf{99.67} &
  \textbf{99.66} &
  \textbf{99.67}\\
MCKELM (RBF)  \mbox{}\protect\newline (32-subsets) (AF) &
  \textbf{99.66} &
  \textbf{99.67} &
  \textbf{99.66} &
  \textbf{99.67}\\
MCKELM (RBF) \mbox{}\protect\newline (16-subsets) (SHAP) &
  \textbf{99.66} &
  \textbf{99.64} &
  \textbf{99.66} &
  \textbf{99.65}\\
MCKELM (RBF) \mbox{}\protect\newline  (32-subsets) (SHAP) &
  \textbf{99.65} &
  \textbf{99.63} &
  \textbf{99.64} &
  \textbf{99.66}\\
MCKELM (Chi-Square)   \mbox{}\protect\newline (16-subsets) (SHAP) &
  \textbf{99.66} &
  \textbf{99.65} &
  \textbf{99.66} &
  \textbf{99.66}\\
MCKELM (Chi-Square) \mbox{}\protect\newline (32-subsets) (SHAP) \mbox{}\protect\newline  &
  \textbf{99.66} &
  \textbf{99.64} &
  \textbf{99.66} &
  \textbf{99.66}\\
\hline 
\end{tabulary}\par 
\end{table*}

% \bgroup
% \fixFloatSize{images/9a78f5a2-6c71-42cf-819a-8f52573f18ea-u280c06ce-074e-4f8a-9e2e-1a6d15fe0f34.png}
% \begin{figure}[!ht]
% \centering \makeatletter\IfFileExists{images/image.png}{\includegraphics{images/image.png}}{}
% \makeatother 
% \caption{{Confusion matrix of linear fully connected classifier in MobileNetV3 and MCKELM at 16 subsers and 32 subsers for food/non-food classification}}
% \label{f-593489c3aa3c}
% \end{figure}
% \egroup

% \bgroup
% \fixFloatSize{images/timecomparisonmckelm.png}
% \begin{figure*}[!htbp]
% \centering \makeatletter\IfFileExists{images/timecomparisonmckelm.png}{\includegraphics{images/timecomparisonmckelm.png}}{}
% \makeatother 
% \caption{{Comparison of training and testing time at different subsers of MCKELM}}
% \label{f-1151c58b60d6}
% \end{figure*}
% \egroup

\section{Discussion}
We state the following considerations based on the experiments carried out for this research work. Existing datasets of food detection including Food5k, are mid-size datasets, and models trained on these datasets are not generalized enough to tackle a wide array of food detection challenges in a real-world environment. For this purpose, at first, we prepared a large dataset for the food detection problem and achieved this objective by using nine publically available datasets and then divided them into two categories: Food and non-food. Following that, we have proposed a novel framework that extracts features from the efficient neural network, employs SHAP values for reliable feature selection, and then multicolumn kernel extreme learning machine for highly accurate food detection.

 For feature extraction, we have explored the potential of efficient neural networks due to their fast inference. Firstly, we used transfer learning on a pre-trained model to fine-tune variants of efficient neural networks for the food detection problem. Following that, our analysis in the results section shows MobileNetV3 has superior performance as it uses the squeeze and excitations strategy giving unequal weights to different channels in contrast to normal weights, hence leading to the superior performance of MobileNetV3, compared to other models for feature extraction in our framework.   
\mbox{}\protect\newline In the second step of our pipeline, we have selected the most reliable features for real-time classification. The attributes from the deep model have high dimensions, which increases the training and testing time. Shapley-value-based explanations showed that all attributes of feature vectors do not contribute equally towards the correct class. We proposed a strategy that selects the top features with the highest Shapley score for the prediction. Table~\ref{tw-3a435ea43fd3} shows that feature selection significantly reduces the training time while having competitive performance.
The advantage of this method over popularly employed RELIEF F is that  Shapley-value-based explanations help us to visualize and understand the regions in the image based on which classifier has made the decision. 

 \mbox{}\protect\newline Finally, in the last step, we solved the challenge faced by a linear-fully connected layer and employed MCKELM to discriminate linearly inseparable features for highly accurate food detection. MCKELM has solved two major problems faced by KELM, especially in the scenario of large datasets. 1) Curse of dimensionality 2) Strongly coupled network structure. Our results on a large food/non-food dataset show that the proposed framework improves performance by 0.93\%, in contrast to MobileNetV3. The improvement is significant as the dataset is large, and the proposed framework classifies \textbf{513} images correctly as compared to MobileNetV3. Similarly, the comparison with other classifiers in our framework also shows the superior performance of MCKELM in terms of accuracy, precision, recall, and F1 score. 
    
\section{Conclusion}
Food detection is a performance-critical task and requires high classification accuracy. This work has proposed a hybrid framework that takes optimal features from the deep model and uses a separate nonlinear classifier after successfully employing the efficient neural network MobileNetV3 for fast feature extraction. There on, select the optimal subset of features by using SHAP values from the gradient explainer. As Gradient Explainer has explained the contribution of the attributes towards the correct class. This approach has improved the interpretability of the feature selection process. Finally and importantly, we have evolved KELM to MCKELM for the classification stage of our pipeline. KELM is unsuitable for large datasets as increasing the number of hidden nodes has also increased the complexity of inner matrix computations. Moreover, KELM has a strongly coupled network structure which makes it inappropriate for a distributed environment in the cloud. Our proposed MCKLEM for the classification has addressed these two problems by employing a multicolumn approach after successfully using the k-d tree to divide the data into N equal subsets. Then, it has trained separate KELM on each subset of data. During the testing process, we selected only the top k nearest subsets instead of the whole network. The comparison on a large food/non-food dataset prepared by us has demonstrated that the proposed algorithm has superior performance than the linear fully-connected classifier in the deep model and other variants of ELM. The multicolumn approach significantly has reduced the computational time by successfully decoupling the network structure. 

\section*{Author Contributions}
Ghalib Ahmed Tahir was responsible for writing algorithms, analysis, writing the article and literature search, and approved the final version as submitted.  Loo Chu Kiong contributed to the study design, reviewed the study for intellectual content, and confirmed the final version as submitted.

\section*{Acknowledgements}
This research was supported by the Universiti Malaya Impact-oriented Interdisciplinary Research Grant Programme (IIRG) - IIRG002C-19HWB,  Universiti Malaya Covid-19 Related Special Research Grant (UMCSRG) CSRG008-2020ST and Partnership Grant (RK012-2019) from University of Malaya.

\section{Conflict of Interest}
The authors declare there is no conflict of interest.

\bibliography{article}

\end{document}